\documentclass[10pt,final,journal]{IEEEtran}

\usepackage{amsmath,amssymb,amsfonts}
\usepackage{graphicx}
\usepackage{textcomp}
\usepackage{algorithm}
\usepackage[noend]{algpseudocode}
\usepackage{setspace}
\usepackage{enumitem}

\usepackage{tabularx}
\usepackage{multirow}
\usepackage{multicol}
\usepackage{diagbox}
\usepackage{adjustbox}

\usepackage{caption}
\usepackage{subcaption}
\usepackage{comment}

\usepackage{xcolor}

\usepackage[numbers,sort&compress]{natbib}

\begin{document}

\title{A Hybrid Training-time and Run-time Defense Against Adversarial Attacks in Modulation Classification}
\author{Lu Zhang$^{1,2}$, Sangarapillai Lambotharan$^{2}$, Gan Zheng$^{2}$, Guisheng Liao$^{1}$, Ambra Demontis$^{3}$, and Fabio Roli$^{4}$\\$^{1}$School of Electronic Engineering, Xidian University,
Xi’an 710071, People’s Republic of China\\$^{2}$Wolfson School of Mechanical, Electrical and Manufacturing Engineering, Loughborough University, Loughborough, UK\\$^{3}$Dept. of Electrical and Electronic Engineering, University of Cagliari, Piazza d’Armi, 09123 Cagliari, Italy\\$^{4}$Department of Informatics, Bioengineering,
Robotics, and Systems Engineering, University of Genova, 16145 Genoa, Italy}

\IEEEtitleabstractindextext{
\begin{abstract}
Motivated by the superior performance of deep learning in many applications including computer vision and natural language processing, several recent studies have focused on applying deep neural network for devising future generations of wireless networks. However, several recent works have pointed out that imperceptible and carefully designed adversarial examples (attacks) can significantly deteriorate the classification accuracy. In this paper, we investigate a defense mechanism based on both training-time and run-time defense techniques for protecting machine learning-based radio signal (modulation) classification against adversarial attacks. The training-time defense consists of adversarial training and label smoothing, while the run-time defense employs a support vector machine-based neural rejection (NR). Considering a white-box scenario and real datasets, we demonstrate that our proposed techniques outperform existing state-of-the-art technologies.
\end{abstract}

\begin{IEEEkeywords}
DNNs, adversarial examples, projected gradient descent algorithm, adversarial training, label smoothing, neural rejection
\end{IEEEkeywords}}

\maketitle

\IEEEdisplaynontitleabstractindextext
\IEEEpeerreviewmaketitle

\section{Introduction}
\IEEEPARstart{I}{n} recent years, deep learning (DL) has prompted significant interests in wireless communications. For example, various researchers \cite{o2018over, scholl2019classification} have successfully applied DL into automatic modulation classification (AMC), which is critical in signal intelligence and surveillance applications including cognitive radio and dynamic spectrum access to monitor spectrum occupancy. Traditionally, the AMC has been achieved by higher-order statistical methods as well as by calculating the compact decision boundary using low-dimensional features that are carefully crafted by the developers. However, AMC can also be implemented by training a deep neural network (DNN) with raw signal samples which allows obtaining highly accurate classifications \cite{ramjee2019fast}. The goal of DL-based AMC is to recognize different types of modulations. The input to the DNN is the raw radio frequency (RF) signals that consist of both the in-phase and quadrature components (IQ samples), as in RML2016.10a \cite{o2016radio}. Taking this RF signal as input, DNN is expected to output probability score for each possible modulation type BPSK, QPSK, 8PSK, QAM16, QAM64, CPFSK, GFSK, PAM4, WBFM, AM-SSB, and AM-DSB. The modulation type is determined by the DNN as the one that produces the largest probability score.

\begin{figure}[ht]
\centering
\includegraphics[scale = 0.4]{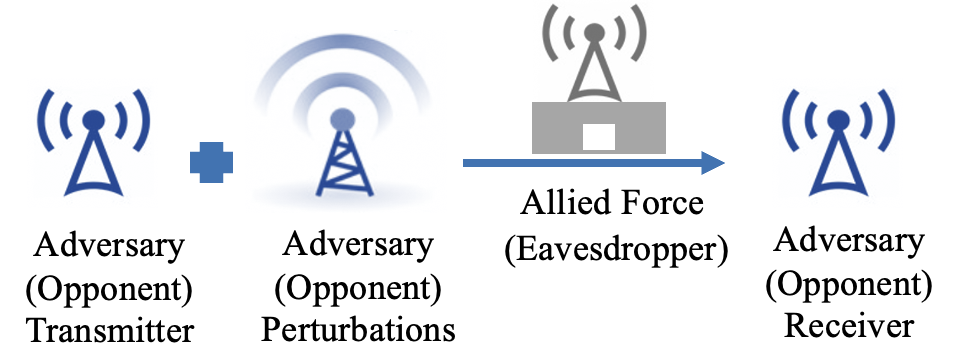}
  \caption{A military scenario for the adversarial examples in modulation classification.}
  \label{fig:eavesdropper}
\end{figure}

Despite their excellent performance, recent studies discovered that DNNs are vulnerable to adversarial examples, i.e., imperceptible and carefully designed modifications of the input that lead to misclassifications \cite{Goodfellow2015}. The earlier works about the vulnerability of DNNs to adversarial examples focus on the computer vision domain. However, Sadeghi et al. \cite{sadeghi2018adversarial} showed that adversarial examples significantly decrease the classification accuracy also in AMC. Notably, AMC is also applied to critical applications such as military scenarios. During warfares, the units of each adversary (opponent transmitter and receiver as shown in Figure \ref{fig:eavesdropper}) exchange crucial information using radio signals. The allied forces (in this scenario taking the role of eavesdropper) can use AMC to discover the modulation used to intercept messages exchanged between the adversarial units (opponents). In order to deter the allied forces to eavesdrop messages, the adversary units (opponent) can make this modulation discovery more difficult by applying to the communication signals small perturbations, crafted ad-hoc to make the automatic discovery of the modulation eventually performed by the allied forces fail (adversarial examples). The allied forces can still discover the modulation, but in this case, it should have an AMC system that is robust to adversarial examples as considered in this paper.

\setlength{\textfloatsep}{0.6mm}

There are only few works investigating the defense against adversarial examples in AMC. For example, Sahay et al. \cite{sahay2021deep} proposed a deep ensemble defense, an ensemble of different deep learning architectures trained on time and frequency domain representations of the received signals. However, this deep ensemble defense is designed for a black box scenario and investigates the transferability between signal domains which is different from a white box scenario as considered in this paper. In \cite{sahay2021robust}, a two-fold defense mechanism, adversarial re-training in conjunction with autoencoder (AE) based detection was proposed. In \cite{kokalj2019mitigation}, an AE-based pre-training was proposed, which trains the encoder of the AE using the same architecture as the targeted CNN and then transfers the weights to the CNN classifier. In \cite{sahay2021robust}, a CNN is used for modulation classifications. The adversary is able to generate adversarial examples based on  the knowledge of the CNN architecture used. However, in order to detect adversarial examples, an AE-based anomaly detector is used by the defense mechanism in \cite{sahay2021robust} (to reject adversarial examples). As our proposed hybrid training-time and run-time defense (HTRD) also works based on the principle of anomaly detection which employs SVM to reject adversarial examples, we consider the work in \cite{sahay2021robust} the most relevant one for performance comparison. Since the generation of adversarial examples in \cite{sahay2021robust} does not consider the AE-based defense, it should be considered as a grey-box attack. Therefore, we have extended the attack scenario in \cite{sahay2021robust} and developed a white-box scenario-based adversarial samples generation using the knowledge of both the CNN and the AE which is compared to our HTRD. More details are given in Section \uppercase\expandafter{\romannumeral3}. In terms of our previous work in \cite{zhang2021countermeasures}, a neural rejection (NR) system and its enhancement, which uses label smoothing and Gaussian noise augmentation (LS-GNA), was proposed. However, as Gaussian noise augmentation resembles a white noise-like attack scenario and ignores the inherent features of the CNN architecture, we adopt a stronger form of attack generated by customized adversarial training (CAT) technique \cite{cheng2020cat}. Briefly, CAT generates adversarial perturbations for each sample based on the parameters and architecture of CNN and customizes the perturbation level and the corresponding label during the adversarial training procedure. Our proposed CAT based technique outperforms the LS-GNA based DNN and the work in \cite{sahay2021robust} as demonstrated using the real radio datasets RML2016.10a \cite{o2016radio}. The motivation behind the proposed HTRD, which combines the CAT technique and NR system, is as follows. Adversarial training has the potential to maximize the input space margin, and the CAT could increase the lower bound of the margin in the input space. Obtaining a DNN with a larger margin in the input space increases the performance of the run-time defense that we consider in this work which is an SVM-based NR detector. This detector can classify those samples that have low confidence as adversarial examples. The adversarial examples are pretty near in input space to their original counterpart. If the classifier has a higher margin, it is less likely to misclassify adversarial examples with high confidence. The attacker has to apply higher perturbations to have them misclassified with the same level of confidence, hence more transmission power is needed to succeed with the modulation classification attacks, which will hinder stealth operation of adversarial transmitter.

\section{The Proposed Countermeasures against Adversarial Attacks}
\subsection{Customized Adversarial Training based DNN}
We propose to employ a robust defense called CAT \cite{cheng2020cat} for modulation classifications, which is a better defense against adversarial examples in AMC as compared to LS-GNA technique. Before delving into details, we first introduce the basics of adversarial training \cite{madry2017towards}, which can be formulated as a min-max optimization. For a K-class classification problem, we denote $D=\left \{(x_{i},y_{i}) \right \}_{i=1,...,n}$ as the set of the training samples with $x_{i}\in\mathbb{R}^{d}, ~y_{i}\in \left \{ 1,...,K \right \}=:[K]$. Let $f_{\theta }(x):\mathbb{R}^{d}\rightarrow [K]$ denote a classification model parameterized by $\theta$. The idea of adversarial training is to generate adversarial examples in each iteration of the training process and add them to the training dataset. Formally, adversarial training can be illustrated as: $\min_{\theta }\frac{1}{n}\sum_{i=1}^{n}\max_{x'_{i}\in \ss(x_{i},\epsilon )}l(f_{\theta }(x'_{i}),y_{i})$,
where $\ss(x_{i},\epsilon )$ is the $l_{p}$-norm ball centred at $x_{i}$ with radius $\epsilon$, and $x'_{i}$ indicates the adversarial examples. The inner optimization problem aims to maximize the loss function between the output of the neural network $f_{\theta }(x'_{i})$ and the true label $y_{i}$ so as to obtain an adversarial example for each data sample $x_{i}$. For a DNN, the inner maximization problem does not have a closed-form solution, hence, normally a gradient-based iterative solver is adopted. In this work, we use the PGD attack \cite{madry2017towards} to solve the inner maximization problem.

However, it is discovered in \cite{cheng2020cat} that a uniformly large $\epsilon$ is often harmful to the performance of the adversarial training because the classifier cannot correctly fit both the training data and the adversarial samples generated in this way. Therefore, the classifier will sacrifice its accuracy on some of the original training samples, which causes a distorted decision boundary. Two useful practices to mitigate this problem have been presented in \cite{cheng2020cat}. First, identical and large $\epsilon$ for all data samples is avoided. Instead, for samples that are originally closer to the decision boundary, a smaller $\epsilon$ is used. This is to ensure that the original normal samples that are close to the decision boundaries do not influence significantly to alter the decision boundary, thereby, not adversely affecting accuracy of the normal data. Second, since a sample with a large perturbation introduces uncertainty on the class it belongs to, label smoothing should be applied instead of one-hot encoding. More label smoothing is applied to samples with higher perturbations.

To address these issues, CAT modifies adversarial training by adapting $\epsilon$ that results in adaptive label smoothing for each training sample. We denote $\epsilon_{i}$ as the level of perturbation allocated to each sample, which is calculated as, $\epsilon_{i} = \mathop{\arg\min}_{\epsilon} \left \{ \max_{x'_{i}\in \ss (x_{i}, \epsilon)} f_{\theta}(x'_{i})\neq y_{i} \right \}$.
It means the $\epsilon$ will not be increased further if the adversarial sample can be misclassified within $\epsilon$ region. For each sample $x_{i}$, the perturbation power is calculated by $\epsilon_{i}^{2}$. On the other hand, CAT uses an adaptive label smoothing method which assigns an adaptively smoothed label for each training sample to accommodate a different perturbation tolerance level to each sample. Specifically, given a one-hot encoded label $y$, the smoothed label is calculated as: $\tilde{y}=(1-\alpha)y+\alpha u$, where $u$ is a uniformly distributed random variable and $\alpha\in [0,1]$ controls the smoothing level. For the adaptive setup, $\alpha=c\epsilon_{i}$ is used so that a larger perturbation tolerance would have a higher label uncertainty and $c$ is a constant. Therefore, the adaptive label is obtained as: 
\begin{equation}
\label{equ:adaptive label smoothing}
\tilde{y_{i}}=(1-c \epsilon_{i})y_{i}+ c \epsilon_{i} u.
\end{equation}
Overall, the objective function of the CAT algorithm is formulated as \cite{cheng2020cat}:
\begin{equation}
\label{equ:objective function of CAT}
\begin{aligned}
&\min_{\theta}\frac{1}{n}\sum_{i=1}^{n}\max_{x'_{i}\in \ss (x_{i},\epsilon_{i})}l(f_{\theta}(x'_{i}),\tilde{y_{i}})\\
&s.t. ~~\epsilon_{i} = \mathop{\arg\min}_{\epsilon} \left \{ \max_{x'_{i}\in \ss (x_{i}, \epsilon)} f_{\theta}(x'_{i})\neq y_{i} \right \}
\end{aligned}
\end{equation}

The algorithm for CAT used for modulation classification is adopted from \cite{cheng2020cat}. The difference is that instead of using PGD attack with $l_{\infty}$-norm constraint, we use PGD attack with the $l_{2}$-norm constraint as $l_{2}$-norm is a natural choice for AMC since it represents perturbation power. Specifically, during each iteration of the adversarial training, for each training sample, an adversarial example is generated attacking the updated model with the PGD algorithm \cite{madry2017towards}. 
After calculating PGD attack for each training sample, the adaptive smoothed label is calculated using \eqref{equ:adaptive label smoothing}. The DNN model is updated based on the calculated PGD attack and its corresponding adaptive smoothed label, during which $\epsilon_{i}$ is limited to make sure that the adversarial perturbation is not too large. Finally, if the generated adversarial example $x'_{i}$ is successful under the current $\epsilon_{i}$, i.e., $f_{\theta}(x'_{i})\neq y_{i}$, we keep the same $\epsilon_{i}$ for this sample in the next training, otherwise $\epsilon_{i}$ will be increased in the subsequent iteration.

\subsection{Insights into the Proposed Hybrid Training-time and Run-time Defense}

\begin{figure}[h]
\centering
\begin{subfigure}{.23\textwidth}
\centering
\includegraphics[scale = 0.33]{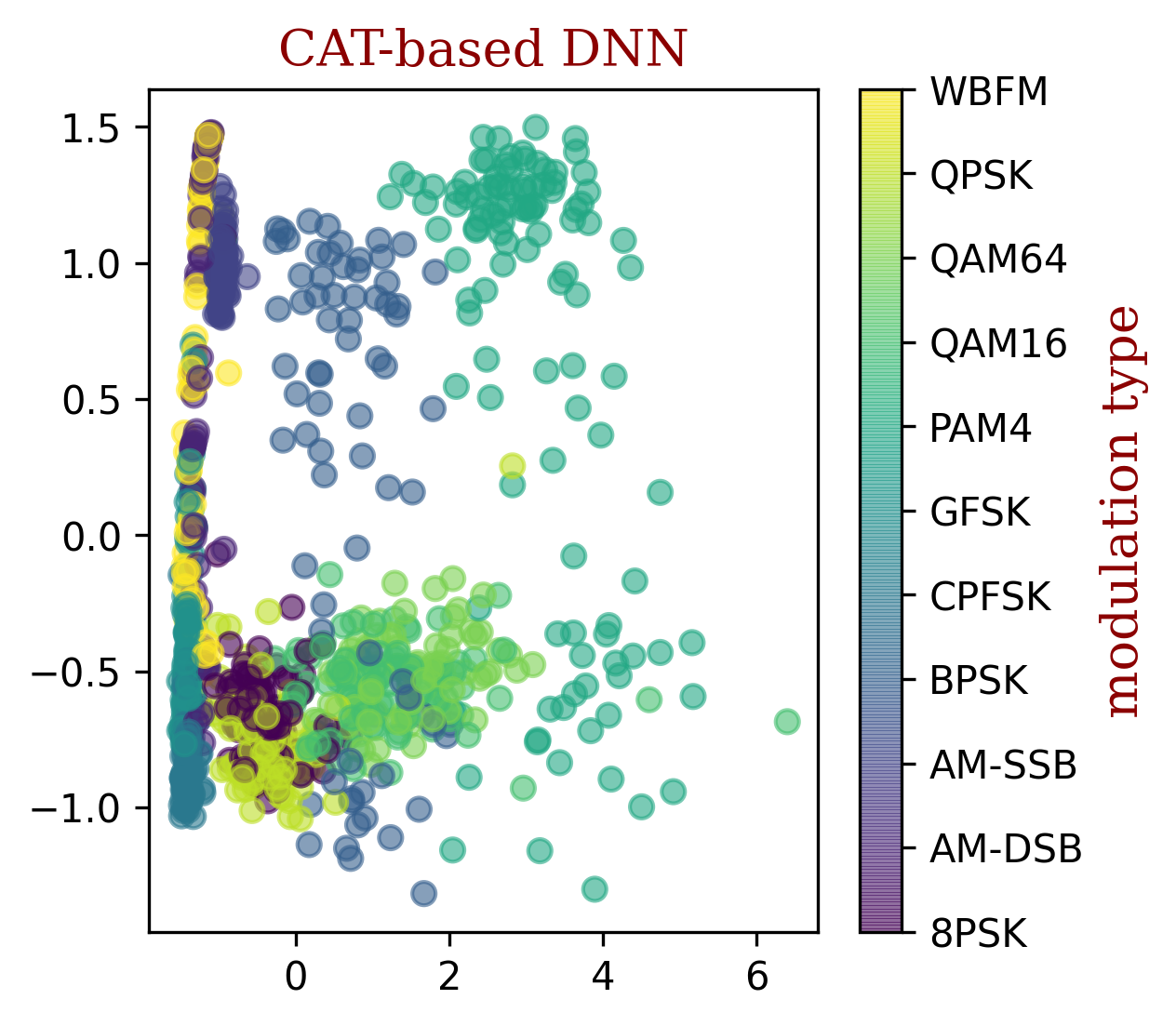}
  \caption{Visualization of the last feature layer of the CAT-based DNN.}
  \label{fig:visual1}
\end{subfigure}
\hfill
\begin{subfigure}{.23\textwidth}
\centering
\includegraphics[scale = 0.33]{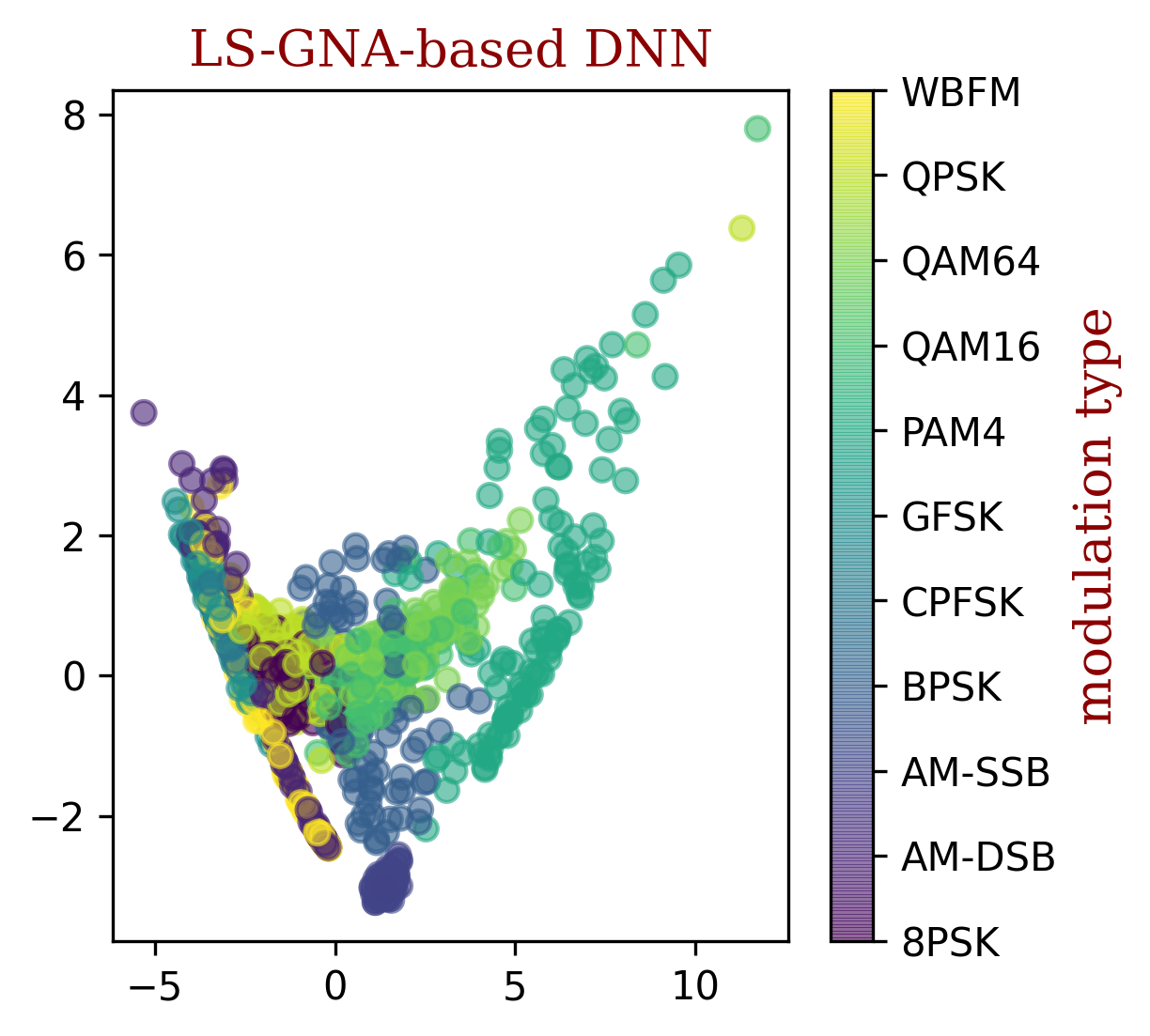}
  \caption{Visualization of the last feature layer of the LS-GNA-based DNN.}
  \label{fig:visual2}
\end{subfigure}
\caption{Visualization of the last feature layer of the DNN based on CAT and LS-GNA techniques.}
\end{figure}

From the input space margin maximization perspective, adversarial training with cross-entropy loss approximately maximizes a lower bound of the margin if the perturbation size $\epsilon $ is not greater than the soft logit margin \cite{ding2018mma}. Therefore, as suggested in \cite{ding2018mma}, for adversarial training it is beneficial to start with a smaller $\epsilon $ and then to increase it gradually during the training as the lower bound of the margin is maximized at the beginning. The CAT technique considered for modulation classification complies with this finding, therefore, the margin of the last feature layer of the CAT-based DNN (i.e., the input space for the connected SVM) is expected to increase. To visualize this, the principal component analysis is used to reduce the dimension, and the visualization of the last feature layer for both the CAT-based DNN and the LS-GNA-based DNN is shown in Figure \ref{fig:visual1} and Figure \ref{fig:visual2}, respectively. It can be seen that the classes are more separated for the last feature layer of the CAT-based DNN. This higher separation will produce a larger rejection region for the NR system. A larger rejection region will force the adversary to use more transmission power to attack the modulation classification scheme, which will hinder stealth operation of the adversarial transmitter.

\subsection{Hybrid Training-time and Run-time Defense}
\begin{figure}[ht]
\centering
\includegraphics[scale = 0.23]{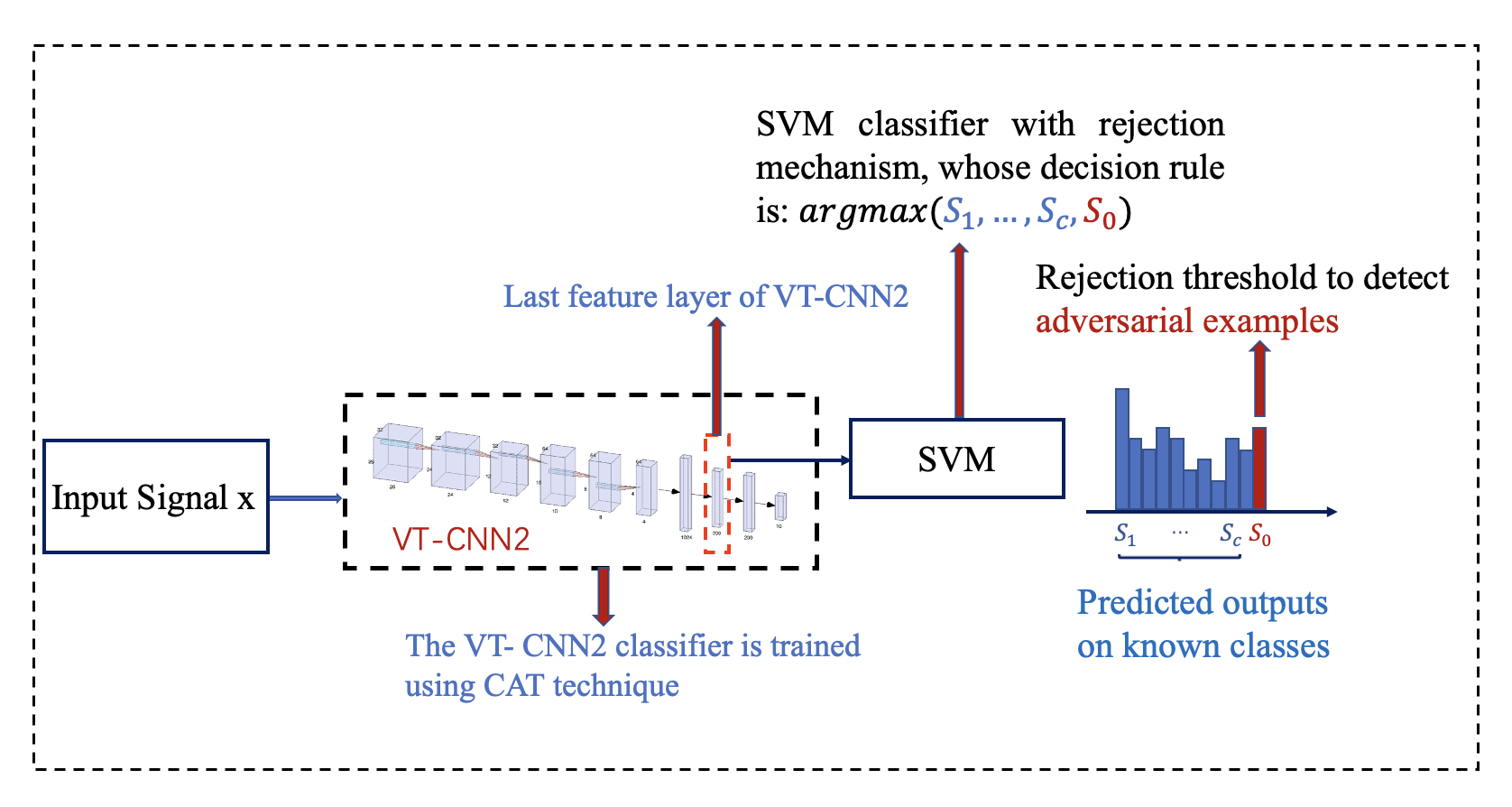}
  \caption{The architecture of the proposed HTRD.} 
  \label{fig:NR}
\end{figure}

Melis et al. \cite{Melis2017} proposed a neural rejection (NR) system to detect adversarial examples. 
The NR system works by establishing a rejection region which could reject the samples that fall within a region of low confidence. The rationale behind the NR system is that there is an amplification effect for the difference between adversarial attacks and benign samples during the propagation of adversarial perturbations through layers of DNN, and this amplification phenomenon is significant at the last layer, hence the NR technique works using the last feature layer of DNN as the features of another classifier, i.e., SVM. The architecture of the proposed HTRD, which combines the benefits of CAT and NR system, is shown in Figure \ref{fig:NR}. The architecture is the same as that in \cite{Melis2017}, however, the difference is that DNN was trained using the CAT technique; hence, the DNN parameters would differ from \cite{Melis2017}. To train the proposed HTRD, we first train the DNN using the CAT technique, then the NR system is employed on the CAT-based DNN. Specifically, the representations $\zeta$ extracted from the last feature layer of the CAT-based DNN are used to train the SVM classifier $S(\cdot)$. Our SVM classifier is a one-vs-all classifier with a radial basis function (RBF)-kernel, and its decision scores (i.e., the output of the SVM) can be presented as: 
\begin{equation}
\label{equ:decision function of SVM}
S(x)=\sum_{i=1}^{K}\alpha _{i}y_{i}\exp(-\gamma \left \| x-x_{i} \right \|^{2})+b
\end{equation}

where $K$ is the total number of support vectors, $\alpha _{i}$ are dual variables, $\gamma$ is the RBF kernel parameter $x_{i}$ is the support vector and $y_{i}$ is the class label for each sample $x'_{i}$. In order to train an RBF-SVM classifier, the hyper-parameter of SVM, $\gamma$, needs to be pre-set. In this work, we use $\gamma=0.01$. During the testing phase, given a new input signal $x$, we take the features $\zeta$ from the pre-trained CAT-based DNN and feed them into the connected RBF-SVM. The decision scores of all the possible classes $S_{1}(\zeta),...,{S_{c}(\zeta)}$ will be obtained from the output of the SVM. The decision function of the HTRD is: 
\begin{equation}
\label{equ:decision_function1}
c^*=\underset{k=1,...,c}{\mathrm{argmax}}\, S_{k}(\zeta), \textup{~only~if~}S_{c^{*}}(\zeta)>S_{0},
\end{equation}
which means the input signal $x$ will only be correctly classified when the maximum of the decision score $S_{c^{*}}$ is greater than a pre-defined threshold $S_{0}$, otherwise $x$ will be classified as an adversarial example and rejected.

We use adaptive (white-box) attacks to evaluate the performance of the proposed HTRD. Consider that the objective function of HTRD is $ \Psi $. Given a data sample $X$ and the maximum size of the adversarial perturbation $\varepsilon$, an attacker with full knowledge of the defense system can perform white-box attacks by solving the following constrained optimization:

\begin{equation}
\begin{aligned}
&~~~~~~~~~x^*=\underset{x^\prime:||x-x^{'}||_{2} \leq\varepsilon}{\mathrm{argmin}}\, \Psi (x^\prime) \\
&\textup{where}  \quad \Psi(x^\prime) = s_{y}(x^\prime) - \max_{j\notin \left \{ 0,y \right \}}s_{j}(x^\prime).
\end{aligned}
\label{equ:attacks_DDNRM}
\end{equation}
In \eqref{equ:attacks_DDNRM}, $||x - x^\prime||_{2}\leq \varepsilon  $ is the $l_{2}$-norm constraint. Besides, $y$ means the true class of the input data sample, and $0$ means the rejection class. Hence, given an input sample $X$, the adversary aims to minimize the confidence score of $X$ that belongs to the true class while maximizing the confidence score of $X$ that belongs to either rejection class or wrong class so as to achieve the untargeted evasion. To solve \eqref{equ:attacks_DDNRM}, the standard PGD algorithm \cite{madry2017towards} was used. Specifically, the gradient of the objective function $\triangledown \Psi (x)$ is first calculated and a standard gradient descent procedure is applied. Then, a projector on the $l_{2}$-norm constraint $||x - x^\prime||_{2}\leq \varepsilon$ is adopted to make the new data point within the feasible constraint domain, where $\varepsilon =\sqrt{PNR\cdot \left \| x \right \|_{2}^{2}/(SNR+1)}$. Here PNR is the perturbation to noise ratio and the SNR is the signal to noise ratio. The iteration will stop when either the objective function converges or the attack can escape the HTRD, i.e., {$||\Psi (x') - \Psi (x)|| \leq t$ or $\max_{j\notin \left \{ 0,y \right \}}s_{j}(x^\prime) > \max_{i\in  \left \{ 0,y \right \}}s_{i}(x^\prime)$}. For our white-box scenario, the attacker has the full knowledge of HTRD achitecture, i.e., the attacker is assumed to have information of both the DNN and the SVM classifier. Specifically, to obtain the gradient $\triangledown \Psi (x)$ for the adversarial examples, we first calculate the gradient of the feature vectors of DNN to the input data $\frac{\partial \zeta}{\partial x}$ using automatic differentiation package and calculate the gradient of RBF-SVM $\triangledown S(\zeta )$ = $\frac{\partial \Psi (x)}{\partial \zeta}$ as $\triangledown S(\zeta )=\sum_{i}^{N}-2\gamma \alpha _{i}y_{i}\exp(-\gamma \left \| \zeta -\zeta_{i} \right \|^{2})*(\zeta-\zeta_{i})$. Finally the overall gradient is obtained using the chain rule as  $\triangledown \Psi (x)=\frac{\partial \Psi (x)}{\partial \zeta}*\frac{\partial \zeta }{\partial x}$.

\section{Results and Discussion}
We evaluate the performance of our proposed CAT-based DNN and HTRD against white-box untargeted adversarial attacks. To compare the effectiveness of the proposed system, we have implemented the state-of-the-art LS-GNA-based NR system \cite{zhang2021countermeasures} and the two-fold defense system proposed in \cite{sahay2021robust}, i.e., adversarial retraining in conjunction with AE detector. To make the comparison fair, we set the relevant thresholds in both the schemes such that  the rejection rate of normal benign samples is equal to 10$\%$ for both schemes. Furthermore, as mentioned before, when generating adversarial examples, the authors of \cite{sahay2021robust} calculate the gradient of the adversarial training considering only the CNN, which is equivalent to a grey-box attack, i.e., the knowledge of the AE-based anomaly detection was not considered. To consider the white-box scenario in \cite{sahay2021robust}, we modified adversarial example generation as follows: We denote the classifier of the adversarial training based CNN and AE detector as $g(\cdot )$ and $h(\cdot )$. For the $k_{th}$ iteration of PGD generation process, we calculate the gradient considering two different conditions. First, if the sample $x_{(k)}'$ could not result in misclassification for the CNN $g(\cdot )$, i.e., $\underset{i}{\mathrm{argmax}}\, g_{i} (x_{(k)}')= true~ class ~y $, the gradient of the PGD is generated with the intention of making misclassification, i.e., maximizing the loss function between $g(x_{(k)}')$ and $y$. The loss function could be the cross entropy loss $CE(g (x_{(k)}'), y)$ or the difference between the logits corresponding to the true class and the most competing wrong class, $g_{y}(x_{(k)}')-\max_{j\notin \left \{ y \right \}}g_{j}(x_{(k)}')$. Second, if $x_{(k)}'$ could result in misclassification for the CNN $g(\cdot )$ but can not escape from the detection of AE, i.e., $\underset{i}{\mathrm{argmax}}\, g_{i} (x_{(k)}')\neq y~\textup{and}~\textup{MSE}(h(x_{(k)}'),x_{(k)}')>Threshold$, the gradient of the PGD is generated with the intention of making  anomaly detection at the AE fails. If the $l_{2}$-norm of adversarial perturbation is larger than the predefined bound $\varepsilon$, the sample $x_{(k+1)}'$ will be projected back to the region such that the adversarial perturbation is within the bound. The PGD iteration will continue until the occurrence of both CNN misclassification and detection of an anomaly at the AE (i.e., $\underset{i}{\mathrm{argmax}}\, g_{i} (x_{(k)}')\neq y~\textup{and}~\textup{MSE}(h(x_{(k)}'),x_{(k)}')<Threshold)$, or until the maximum iteration is reached. 

\subsection{Experimental Setup}
\subsubsection{Dataset and Classifier}
The GNU radio ML dataset RML2016.10a \cite{o2016radio} contains 220,000 input samples, representing 11 different modulations BPSK, QPSK, 8PSK, QAM16, QAM64, CPFSK, GFSK, PAM4, WBFM, AM-SSB, and AM-DSB. These samples are crafted using 20 different SNR levels from -20dB to 18dB with a step of 2dB. We use half of the dataset as the training set and the rest as the testing set. The same VT-CNN2 classifier as in \cite{sadeghi2018adversarial} was used.

\subsubsection{Parameter Setting}
For evaluating the security of the proposed defense system, we used 1000 data samples from the test set, which corresponds to $\textup{SNR} = 10\textup{dB}$. Accordingly, a total of 1000 white-box PGD-based attacks were generated for each algorithm. 

\subsubsection{Security Evaluation}
To test the performance, we calculated the accuracy rate against various adversarial perturbations $\varepsilon$ (or PNR). When there is no perturbation, i.e., $\varepsilon = 0$, the rejection rate is the misclassification rate of benign input data, i.e., unperturbed data that are wrongly rejected by the HTRD. For adversarial examples, i.e., $\varepsilon > 0$, the accuracy rate indicates the rate of the adversarial examples that are either rejected by the rejection mechanism or correctly classified.


\subsection{Experimental Results}
\begin{figure}[ht]
\centering
\includegraphics[scale = 0.45]{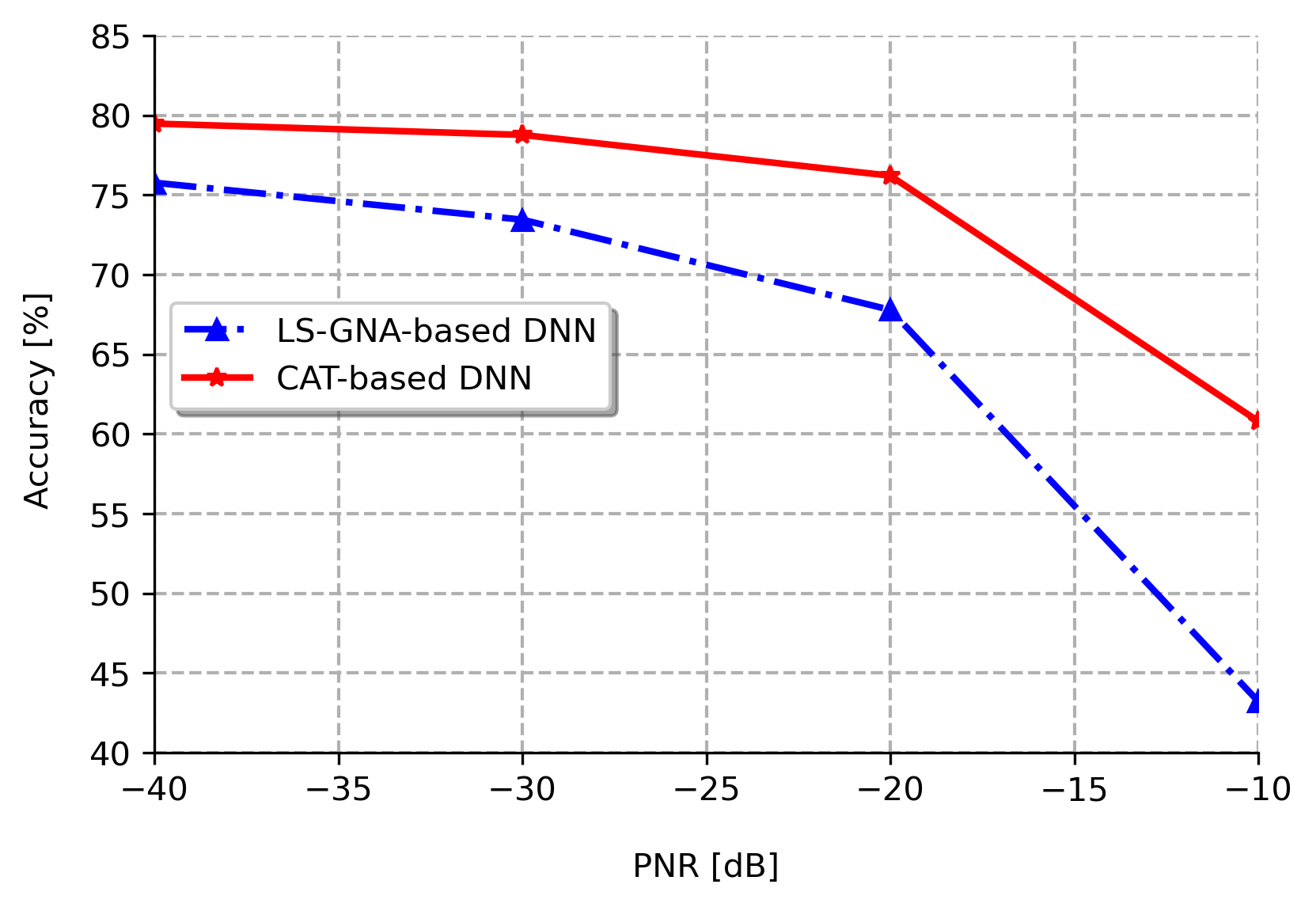}
  \caption{Accuracy of the CAT-based DNN countermeasure as compared to the LS-GNA-based DNN.}
  \label{fig:result1}
\end{figure}

\begin{figure}[ht]
\centering
\includegraphics[scale = 0.45]{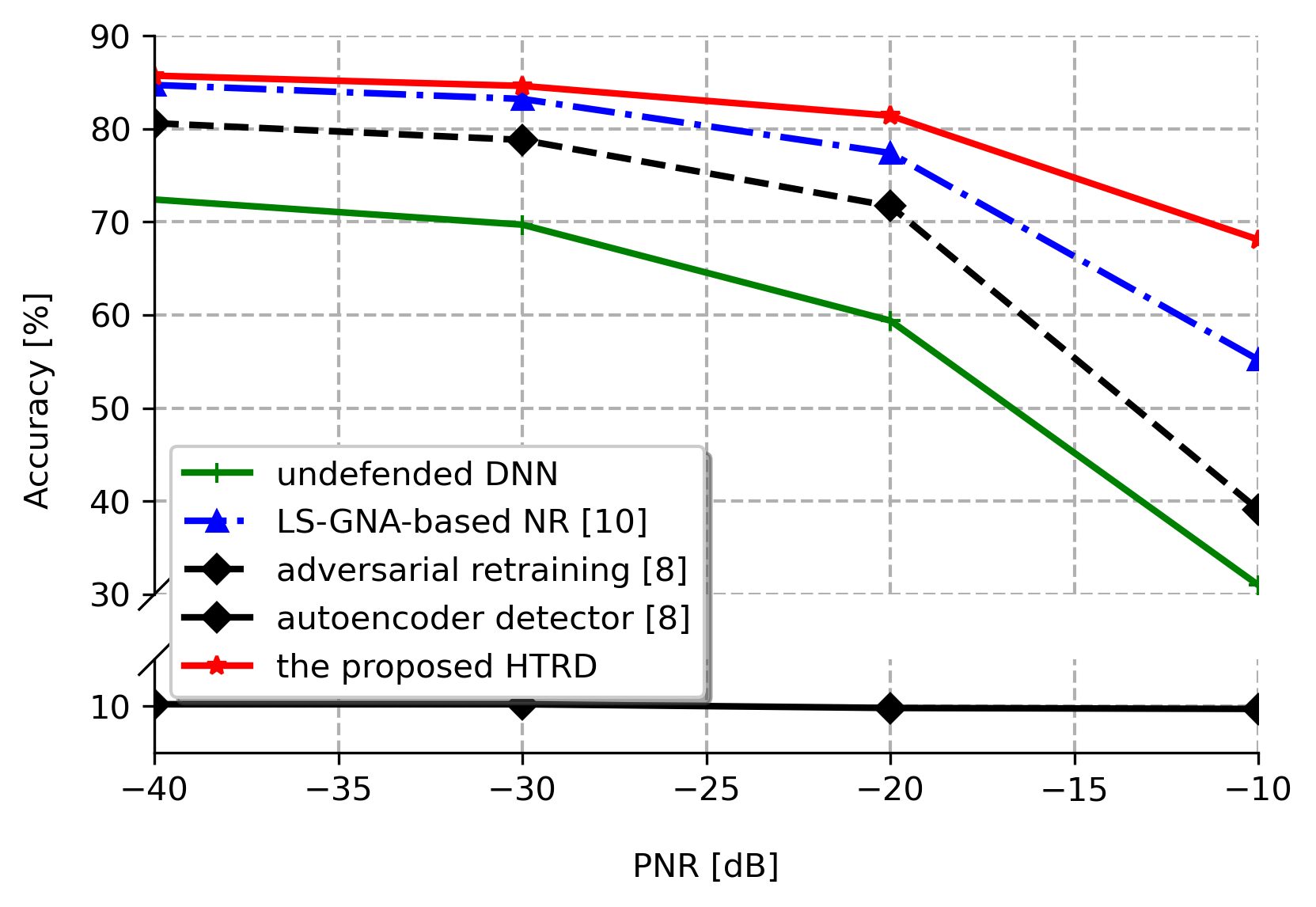}
  \caption{Accuracy of the proposed HTRD countermeasure as compared to the LS-GNA-based NR countermeasure, the two-fold defense \cite{sahay2021robust} and the undefended DNN.}
  \label{fig:result2}
\end{figure}

Figure \ref{fig:result1} depicts the performance of DNN based on two different defense techniques: CAT, and LS-GNA \cite{Shafahi2019}. As seen, the CAT-based DNN achieves higher accuracy than the LS-GNA-based DNN for a wide range of PNR. Especially, the improvement is more significant for higher PNR; for example, when PNR = -10dB, the CAT-based DNN achieves 17$\%$ higher accuracy than the LS-GNA-based DNN. Figure \ref{fig:result2} depicts the performance of countermeasures based on the NR system. Specifically, the proposed HTRD outperforms the state-of-the-art LS-GNA-based NR system \cite{zhang2021countermeasures} and the two-fold defense system \cite{sahay2021robust}. The gap becomes larger as PNR increases; for example, for PNR = -10dB, it improves about 12$\%$ as compared to the LS-GNA based NR system and 30$\%$ as compared to the adversarial re-training based DNN. For the white-box scenario, the accuracy of the auto-encoder-based detector is 10$\%$. We have also observed that the normal accuracy for the undefended DNN is 74.5$\%$, and for our proposed HTRD, the normal accuracy is 75.5$\%$ without NR and 78.7$\%$ with NR. Hence, the proposed HTRD is able to improve the robustness against adversarial examples without sacrificing the normal accuracy.

\section{Conclusion}
We have proposed two defense schemes, namely a CAT-based DNN and a HTRD countermeasure for modulation classifications. Using real radio signals, we have shown that the proposed HTRD scheme, based on the adversarial training, label smoothing and neural rejection outperforms the state-of-the-art LS-GNA-based NR scheme and the two-fold defense mechanism. As a result, the adversary will be forced to use more transmission power to fool the HTRD used by the defender.

\section*{Acknowledgments}
The authors acknowledge the support of Engineering and Physical Sciences Research Council through grants EP/R006385/1 and EP/N007840/1.

\small{
    \bibliographystyle{IEEEtran}
    \bibliography{references}
}

\end{document}